\documentclass[letterpaper, 10 pt, conference]{ieeeconf}
\IEEEoverridecommandlockouts
\usepackage{cite}
\usepackage{amsmath,amssymb,amsfonts}
\usepackage{algorithmic}
\usepackage{graphicx}
\usepackage{textcomp}
\usepackage{xcolor}
\usepackage{multirow}
\usepackage{adjustbox}
\usepackage{booktabs}

\def\BibTeX{{\rm B\kern-.05em{\sc i\kern-.025em b}\kern-.08em
    T\kern-.1667em\lower.7ex\hbox{E}\kern-.125emX}}
\begin{document}
\bstctlcite{MyBSTcontrol}

\title{Autonomous social robot navigation in unknown urban environments using semantic segmentation$^{*}$}

\author{Sophie Buckeridge, Pamela Carreno-Medrano, Akansel Cosgun, Elizabeth Croft, and Wesley P. Chan$^{1}$% <-this % stops a space
\thanks{*This project is supported by the Foundation for Australia-Japan Studies under the Rio Tinto Australia-Japan Collaboration Project.}% <-this % stops a space
\thanks{All authors are with the Faculty of Engineering, Monash University, Melbourne, Australia}%
\thanks{$^{1}$Contact Author
        {\tt\small wesley.chan@monash.edu}}%
}

\maketitle

\begin{abstract}
For autonomous robots navigating in urban environments, it is important for the robot to stay on the designated path of travel (i.e., the footpath or sidewalk), and avoid areas such as grass and garden beds, for safety and social conformity considerations. This paper presents an autonomous navigation approach for unknown urban environments that combines the use of semantic segmentation and LiDAR data. The proposed approach uses the segmented image mask to create a 3D obstacle map of the environment, from which the boundaries of the footpath is computed. Compared to existing methods, our approach does not require a pre-built map and provides a 3D understanding of the safe region of travel, enabling the robot to plan any path through the footpath. Experiments comparing our method with two alternatives using only LiDAR or only semantic segmentation show that overall our proposed approach performs significantly better with greater than 91\% success rate outdoors, and greater than 66\% indoors. Our method enabled the robot to remain on the safe path of travel at all times, and reduced the number of collisions. 
\end{abstract}

\section{Introduction}
Recently there has been increasing interest in the development of last-mile delivery robots. Such robots are envisioned to navigate in urban environments, sharing footpaths with people, and delivering packages to the end customer. This is the most inefficient stage due to the large labour force needed and the increased traffic and parking limitations. Traditional delivery processes require human drivers to operate road vehicles which exacerbates these issues \cite{chen_adoption_2021}. It is expected that autonomous delivery robots can help alleviate these problems.

Package delivery robots must be able to identify safe and appropriate paths in order to autonomously navigate in a manner that is safe and considered to be socially acceptable to other people it shares the path with. A robot should avoid ditches or water puddles for safety reasons, because traversing through these areas may cause damage to the robot. However, a robot should perhaps also avoid grass patches, even though it can safely traverse through it, because people may consider it to be socially inappropriate to trample on the grass, if it were a well-kept lawn.
Existing approaches for safe/appropriate path detection and autonomous navigation often requires the use of a pre-built global map (e.g., \cite{vision-based-road-following, autonomous-city-center, indoor-target-semantic}), or only use 2D images to plan paths on well defined and relatively simple footpaths \cite{robot-monocular-vision}. 
%In a package delivery context, time is a crucial constraint so the time required to build maps is a key limiting factor. Methods that only have 2D information are largely restricted to set target points down the center of the path, making the robot less adaptable to avoid dynamic obstacles (e.g., pedestrians). 
Furthermore, existing approaches (e.g., \cite{vision-based-road-following, learning-based-semantic-navigation, indoor-slam-semantic}) have limited evaluation in real world navigation settings. Without extensive testing it is difficult to gauge how well these methods would perform in various unknown environments.

\begin{figure}[t]
\centerline{\includegraphics[height=80pt]{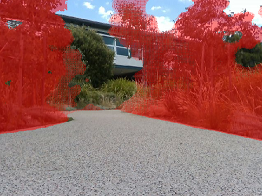} \includegraphics[height=80pt]{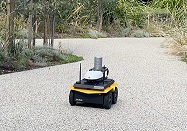}}
\caption{Left image shows segmented point cloud of obstacle environment. Right image shows Jackal robot autonomously navigating within the footpath.}
\label{front-image}
\end{figure}

In this paper, we propose a method that enables a mobile robot to autonomously navigate to an a priori unknown environment while remaining on the footpath and avoiding obstacles (Fig \ref{front-image}). Our approach addresses main drawbacks of existing methods in that we do not require any pre-built maps and we provide a 3D understanding of the safe and socially appropriate region of travel to the robot, enabling less restrictive path planning. Our approach utilizes semantic segmentation and LiDAR data for identifying the boundaries of the safe and appropriate path of travel. We evaluated our approach in three indoor scenarios and four outdoor environments - including challenging environments with paths surrounded by grass or interior glass walls/doors. Results show that our proposed approach yields the best performance with the highest success rate, when compared to alternative methods.

In the following, Section \ref{related-works} reviews the current state of the art. Section \ref{proposed-approach} presents our proposed method. Our experiment setup and results are detailed in Sections \ref{experimental-setup} and \ref{experimental-results}, followed by discussions in Section \ref{discussion}. Finally, Section \ref{conclusion} concludes the paper.

\section{Related Works}
\label{related-works}
A last-mile package delivery robot is required to first detect safe and appropriate paths of travel, and then autonomously navigate along these paths. This section reviews the current state of the art in footpath detection, autonomous navigation and the combination of these two components. 

\subsection{Footpath Detection}
Previous works for footpath detection have included using a camera-based approach \cite{all-day-outdoor-robot, semantic_pothole, robot-monocular-vision, learning-based-semantic-navigation, vision-based-road-following, indoor-target-semantic, semantic_seg_indoor}, a laser sensor based approach \cite{lidar_detection, laser-footpath-damage}, a fusion of the two \cite{real-time-semantic-mapping, lidar-fusion-lane-detection, fusion-transport-robot} or a depth camera-based approach\cite{rgb-d-cnn, rgb-d-semantic-seg, indoor-slam-semantic}. Camera-based approaches are often favored because of their lower cost, many employing a neural network to segment camera images into footpath and obstacles. Semantic segmentation is common  \cite{all-day-outdoor-robot, semantic_pothole, semantic_seg_indoor, real-time-semantic-mapping, vision-based-road-following, learning-based-semantic-navigation, indoor-target-semantic, rgb-d-semantic-seg, rgb-d-cnn, lidar-fusion-lane-detection} with many models using deep convolutional networks (CNNs) \cite{learning-based-semantic-navigation, semantic_pothole, real-time-semantic-mapping, semantic_seg_indoor, rgb-d-cnn, lidar-fusion-lane-detection} due to their high accuracy. A CNN utilizing the U-Net architecture \cite{semantic_pothole} for road condition monitoring produced high level of accuracy (97\%), showing the U-Net architecture's ability to be used in applications other than its original use in medical image segmentation. A limitation of camera-based approaches is the lack of depth information which is important for navigation in 3D environments. Laser sensor-based methods provide 3D information that can better handle subtle variations in path height as well as lighting variations. Road surveying is a common application \cite{laser-footpath-damage}, using LiDAR to identify small variations of the road surface. An approach presented in \cite{lidar_detection} uses LiDAR to detect curbs which dictate the boundaries of footpaths. However, this method is not capable of distinguishing grass patches from footpath. The cost of LiDAR and the sensor's difficulty in detecting glass and other reflective surfaces are key limitations of this type of approaches. Works fusing camera and laser scan sensors aim to handle glass and reflective surface detection through segmentation while also providing 3D information about the environment that is not provided by the 2D camera images. The approach in \cite{lidar-fusion-lane-detection, real-time-semantic-mapping} utilizes an RGB camera to perform semantic segmentation with distance information from LiDAR sensor. The development of RGB-D cameras presents a more cost-efficient means of fusing 2D camera with 3D depth information than expensive LiDAR sensors \cite{rgb-d-semantic-seg, rgb-d-cnn}.

\subsection{Autonomous Navigation}
Autonomous navigation models can be split into two main categories: navigation with a map of the area and navigation without a map. Map-based approaches create more accurate and efficient global routes for the robot to follow. However, in urban environments detailed maps of the safe paths of travel are not easily available. Works exist \cite{autonomous-city-center, indoor-slam-semantic} that utilize simultaneous localisation and mapping (SLAM). In \cite{autonomous-city-center}, unsafe surfaces of travel (e.g. grass) were identified by using remission values returned from the laser scanner combined with inertial measurement unit (IMU) vibrations to assess different types of ground surfaces (e.g. cobblestone pavement). While this work cited that 3 hours to build a map of a 7.4 km path was a reasonable amount of time, this time overhead would be infeasible in a package delivery context where robots would likely be dispatched to different locations each requiring time to build a map before delivery. The SLAM approach is combined with an object detection CNN in \cite{indoor-slam-semantic} to identify obstacles in the environment using an RGB-D camera. This work performed well in simulation but had limited evaluation in real-world navigation.

\subsection{Footpath Detection and Autonomous Navigation}
While there are many works on developing models for footpath detection, there are fewer works incorporating this with navigation that are tested in urban environments. Works that have achieved this fall into two categories: navigation with a map of the environment (e.g., \cite{vision-based-road-following, indoor-target-semantic, autonomous-city-center}) or navigation only using 2D camera information to identify road boundaries and setting target points for the robot to move towards (e.g., \cite{ robot-monocular-vision}). Approaches using a map include the work done in \cite{vision-based-road-following, indoor-target-semantic} where a topological map is used for global path planning and semantic segmentation allows for identification of the movable area locally. Other works use LiDAR to build semantic maps of the environment such as in \cite{real-time-semantic-mapping} which uses LiDAR sensors to obtain geometric information about the terrain and uses semantic segmentation of RGB images for additional information. A limitation of the map-based methods is the additional time required to build the map. The other footpath navigation methods use identified path boundaries to set target points down the center of the path. In \cite{learning-based-semantic-navigation}, a deep CNN model segments the image, processes it to find the trajectory line which is passed to a fuzzy controller. The approach for road following presented in \cite{vision-based-road-following} combines these two types - it utilizes a topological map as well as footpath boundary detection to set central target points. The restriction on the robot's line of movement in these approaches would prove difficult in pedestrian heavy environments where the impact of these dynamic obstacles would increase the difficulty in calculating this target points as highlighted in \cite{vision-based-road-following}.

The proposed approach in this paper aims to address the limitations discussed above. Our approach utilizes an RGB-D camera for semantic segmentation, suitable due to its low cost. Our approach also does not require a pre-built map and determines road boundaries in real time. Our approach utilizes 3D depth information from the RGB-D to convert the segmented mask to a laser scan data  representing the path boundaries, and does not restrict the robot to navigate along the center of the path. 

\section{Proposed Approach}
\label{proposed-approach}

\begin{figure}[t]
\centerline{\includegraphics[width=200pt]{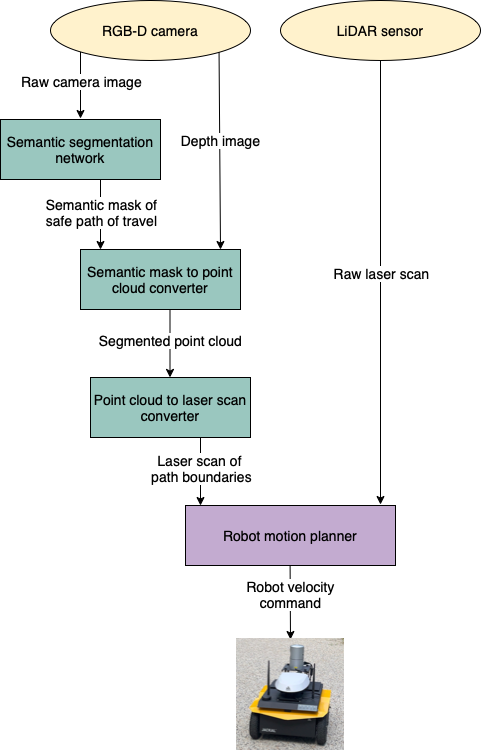}}
\caption{System overview. Yellow ovals represent the sensor hardware. Green boxes represents system components for identifying footpath boundary. Purple boxes represents the path planner and navigation component.}
\label{system overview}
\end{figure}

Fig.~\ref{system overview} illustrates our system and proposed approach. Our system consists of four main components: a semantic segmentation network, a semantic mask to point cloud converter, a point cloud to laser scan converter, and a local robot motion planner. In the following, we detail each component.

\subsubsection{\textbf{Semantic Segmentation Network}}
Given an input RGB image $I$, our semantic segmentation neural network labels each image pixel as either "footpath", $I_{footpath}$, or "background", $I_{background}$. Our semantic segmentation network uses the U-net architecture proposed by Ronneberger et al. \cite{unet} with a Resnet18 \cite{resnet} encoder. The U-net architecture has the benefit of allowing a higher resolution output, whilst requiring fewer training images and producing more precise segmentations. The residal neural network (ResNet) architecture also prevents overfitting and improves precision by adopting average pools as opposed to fully connected layers \cite{Zhang2020}.
These features allow our system to detect the sidewalk region and boundaries with higher accuracy. (For further details of U-net and ResNet, refer to \cite{unet} and \cite{resnet}.) 

\begin{figure}[t]
\vspace{3mm}
\centerline{
\includegraphics[width=78pt]{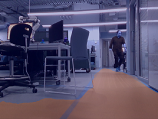} \includegraphics[width=78pt]{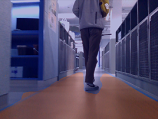} \includegraphics[width=78pt]{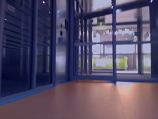}
}

\centerline{
\includegraphics[width=78pt]{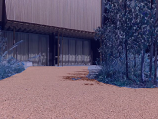} \includegraphics[width=78pt]{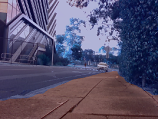} \includegraphics[width=78pt]{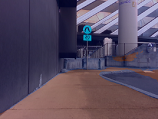}
}

\caption{Example outputs of semantic segmentation model. Top row shows results of indoor model and bottom row shows outdoor model. Pixels labeled as footpath are shaded orange.}
\label{fig:segmentation-results}
\end{figure}

We trained two semantic segmentation models, one for indoor and one for outdoor environments. A total of 200 indoor images of an office-like environment were captured using the camera mounted on our robot while it was driven around the Monash University Robotics Lab. A total of 263 outdoor images were captured while our robot was driven around the Monash University Clayton Campus. We used 30 of the indoor images and 34 of the outdoor images for testing, and the remaining images for training. All images were manually labelled, with each pixel labelled as either "footpath", or "background" for everything else. We augmented the training data by flipping 50\% of the images horizontally.
Fig. \ref{fig:segmentation-results} shows some example results, while Table \ref{semantic-results} show the performance, of our trained models. The two models were trained and tested on a laptop with an Intel Core i7-7700HQ CPU, 16GB of RAM and a Nvidia GTX 1060 6GB.

\begin{table}[]
\caption{Semantic segmentation model test results}
\label{semantic-results}
\begin{tabular}{@{}llll@{}}
\toprule
\multicolumn{2}{l}{}                      & Mean measurement & Standard deviation \\ \midrule
\multirow{2}{*}{Indoor model}  & Accuracy & 96.87 \%         & 2.681 \%           \\
                               & Time     & 30.21 ms         & 2.006 ms           \\
\multirow{2}{*}{Outdoor model} & Accuracy & 97.31 \%         & 1.48 \%            \\
                               & Time     & 22.385 ms        & 1.241 ms            \\ \bottomrule 
\end{tabular}
\end{table}

\subsubsection{\textbf{Semantic Mask to Point Cloud Converter}}
The RGB-D camera outputs an RGB image, $I$, as well as a registered pointcloud $P_{registered}$. The registered pointcloud $P_{registered}$ provides the corresponding 3D points to each image pixel in $I$. Using the semantic segmentation mask of the footpath, along with $P_{registered}$, we generate an obstacle pointcloud, $P_{obstacle}$, by selecting the pointcloud points in $P_{registered}$ corresponding to images pixels in $I$ that are labeled as "background", $I_{background}$ in the segmentation mask. 

\subsubsection{\textbf{Pointcloud to Laser Scan Converter}}
\begin{figure}[b]
\centerline{
\includegraphics[width=95pt]{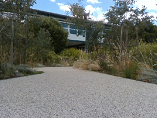} \includegraphics[width=95pt]{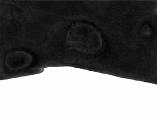}
}
\centerline{
\includegraphics[width=95pt]{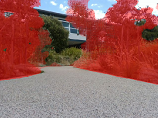} \includegraphics[width=95pt]{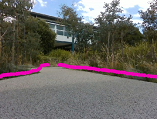}
}
\caption{Top left shows raw camera image. Top right shows output of semantic segmentation network. Bottom left shows the output of semantic mask to pointcloud converter (red points are $P_{obstacle}$). Bottom right shows the output of pointcloud to laser scan converter (pink dots are $L_{obstacle}$).}
\label{model-example}
\end{figure}

The 3D obstacle pointcloud $P_{obstacle}$ is then converted into a 2D laser scan representation $L_{obstacle}$, by flattening the points in $P_{obstacle}$ along the vertical axis \footnote{We used the ROS package pointcloud\_to\_laserscan \cite{pointcloudToLaserscan} in our implementation for this operation.}. This simulated laser scan data then marks the boundaries of the footpath and the edges of safe path of travel to use during navigation (Fig. \ref{model-example}). 

\subsubsection{\textbf{Robot Motion Planner}}
The robot motion planner combines the robot's LiDAR sensor data, as well as $L_{obstacle}$ to build a local costmap around the robot. This costmap is then used to plan a path for the robot to get to the specified goal, while avoiding obstacles and staying on the footpath at the same time. Using the LiDAR data in addition to $L_{obstacle}$ allows the robot to avoid collisions with obstacles even when the semantic segmentation network mislabels an obstacle. It also provides a much wider field of view (360 degrees) compared to the RGB-D camera. We used the move\_base package in the Robot Operating System (ROS) navigation stack as our path planner \cite{ros_navigation}.

\section{Experimental Setup}
\label{experimental-setup}
We evaluated the performance of our proposed approach in a variety of indoor and outdoor environments. We compared our proposed approach with two alternative methods: 1) \textbf{SS Only} which uses only semantic segmentation without LiDAR data, and 2) \textbf{LiDAR Only} which uses only LiDAR data without semantic segmentation. We note that \textbf{LiDAR Only} simply uses the ROS navigation stack and thus acts as a baseline comparison. We tested seven scenarios (four outdoor, and three indoor). In each scenario, we specified a goal point as a coordinate point relative to the robot's starting pose. The robot relies only on wheel encoders for odometry and does not use any maps for localization. 

The experimenter would intervene in order to prevent the robot from colliding into an obstacle that would cause damage. Additional intervention was also required to move the robot past LiDAR sensor noise or mislabelled footpath pixels from the semantic segmentation model when this prevented the Robot Motion Planner from calculating a valid trajectory. The number of times intervention occurred was recorded as assists over the trials.

Each method is tested with 1-2 pedestrians present and without pedestrians. We repeated six trials per location - three trials with pedestrians present and three trials without pedestrians present. Thus, a total of 42 trials per method were conducted for evaluation. 

\subsubsection{\textbf{Hardware}}
We used a Clearpath Robotics Jackal mobile base robot (Fig. \ref{system overview}) equipped with a Velodyne HDL-32 LiDAR sensor and a RealSense RGB-D camera as our robot platform. The Jackal's computer has an Intel I7-8700 CPU, 32GB RAM and Nvidia GTX 1050 GPU with 4GB of GPU memory. The raw camera image of the RealSense has a publish rate of 30 Hz and the depth image at 4.4 Hz. When running our model on the Jackal, the path boundary laser scan result is outputted at a rate of 4.4 Hz, staying in time with the depth image.

\subsubsection{\textbf{Test Environments}}
\begin{figure}[t]
\vspace{3mm}
\centerline{\includegraphics[width=95pt]{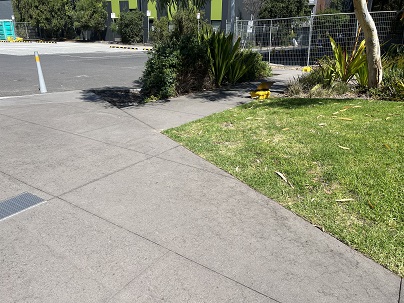}
\includegraphics[width=95pt]{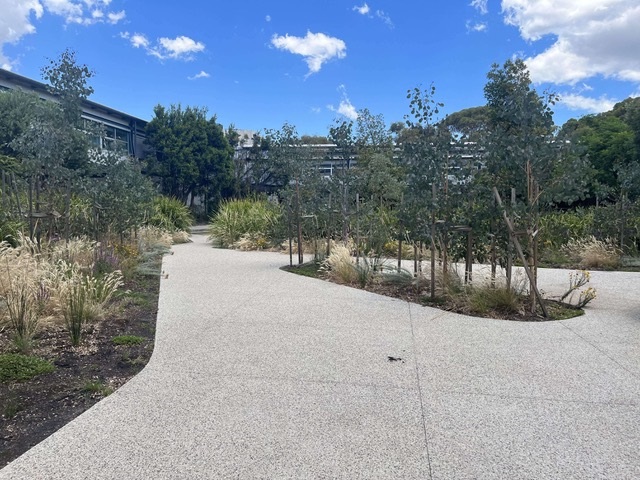}
}
\centerline{\includegraphics[width=95pt]{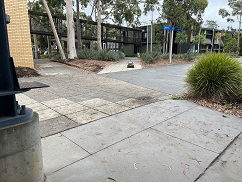}
\includegraphics[width=95pt]{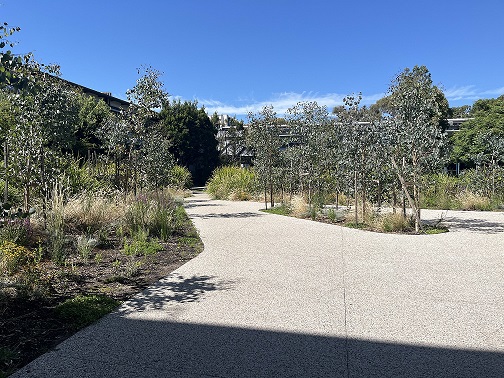}
}
\caption{Outdoor testing locations}
\label{outdoor-environments}
\end{figure}

\begin{figure}[t]
\centerline{\includegraphics[width=95pt]{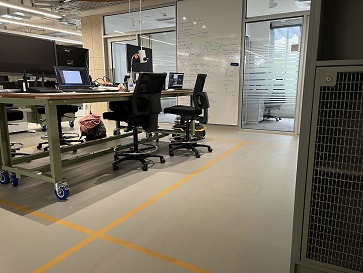}
\includegraphics[width=95pt]{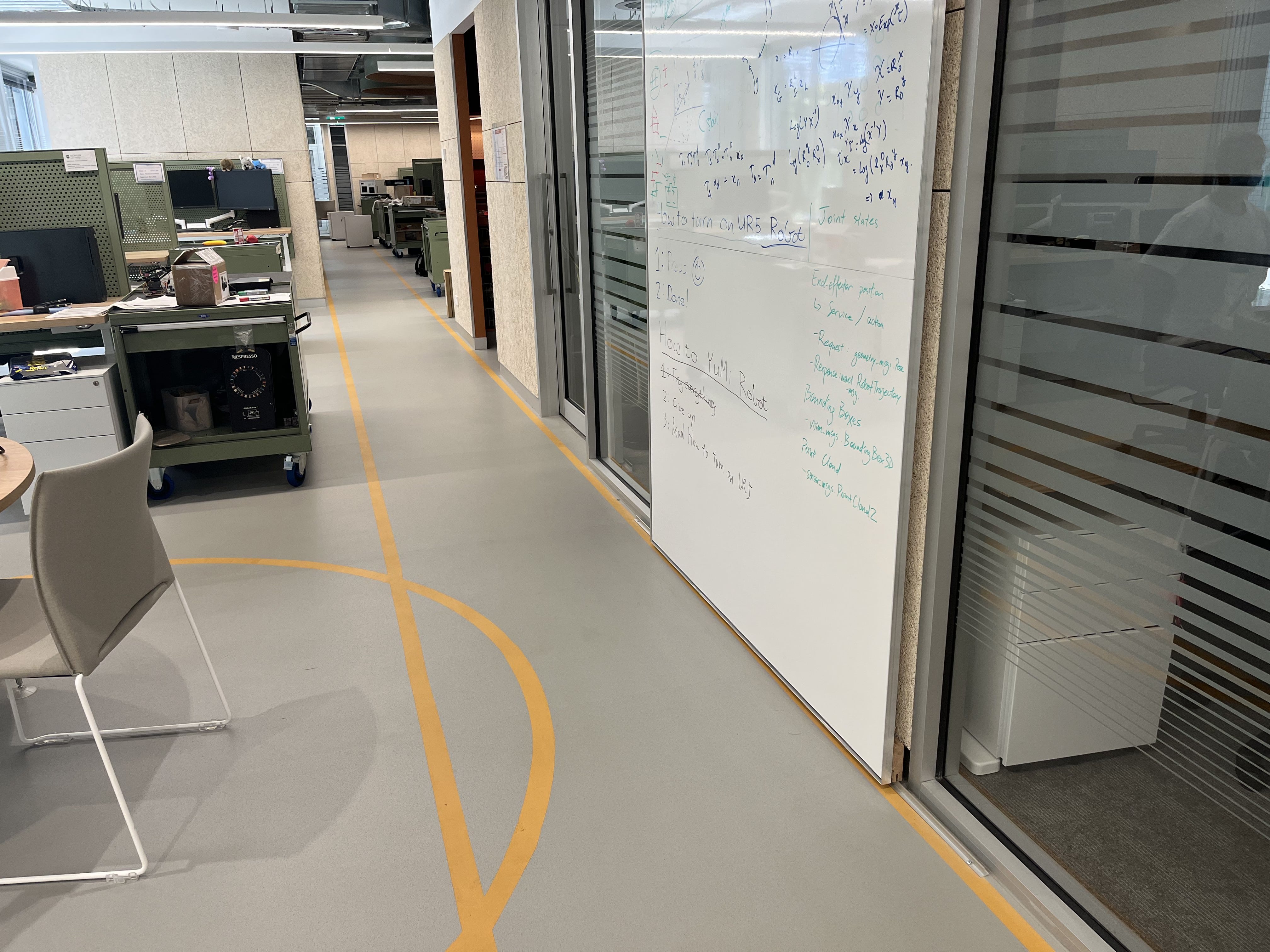}
}
\centerline{\includegraphics[width=95pt]{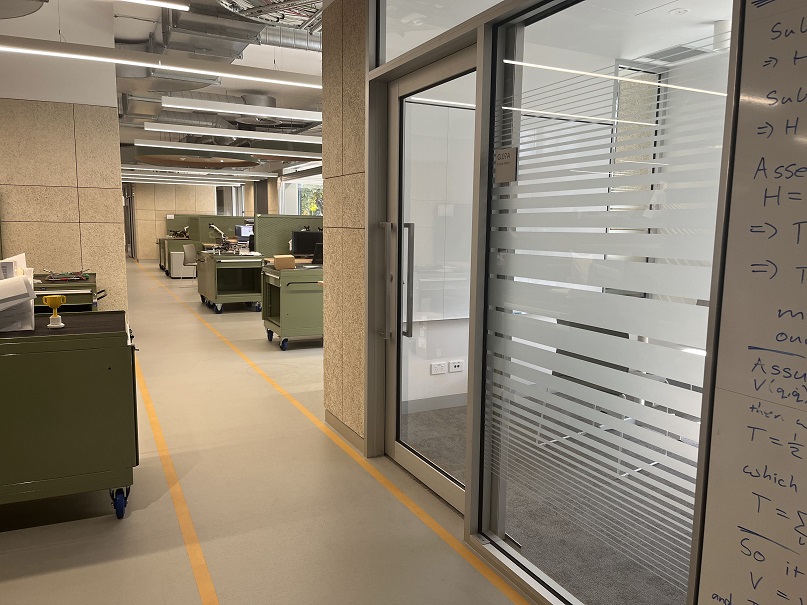}
}
\caption{Indoor testing locations}
\label{indoor-environments}
\end{figure}

The test environments for each navigation scenario were selected from different locations on Monash University Clayton campus, with the aim that each location is representative of a different common type of urban environment. 
Fig.~\ref{outdoor-environments} shows the outdoor environments selected. The top left image location tests if the robot would remain on the footpath to avoid moving on the grass. The top right image shows the location for testing if the robot is able to avoid garden beds that surround the path, and garden bed islands located in the center of a wide footpath area. The bottom left image shows a footpath that is made up of patches of different materials and appearances along the path. Finally, the bottom right image shows a path on a sunny day that is partially covered by shadows, which is usually challenging for image processing.
Fig.~\ref{indoor-environments} shows the indoor testing scenarios. The selected scenarios require the robot to drive down corridors with different widths, around corners, while avoiding pedestrians, tables and chairs. The selected indoor scenarios also included a number of glass doors, which can be very difficult to detect using cameras and LiDARs. 

%To make this experiment feasible, it was assumed that there is a suitable path from the starting point to the goal, making it possible for the robot to reach the goal. It is also assumed that a footpath is characterised as a surface that pedestrians commonly walk along so that the robot will follow the social norms of the surroundings.

\subsubsection{\textbf{Evaluation Metrics}}
We recorded the following metrics for each experiment trial:
\begin{itemize}
  \item[]\textbf{Success of trial}: A trial is considered successful if the robot reaches the goal without collisions, without being assisted and remaining on the footpath.
  \item[] \textbf{Trial duration}: The total time required for the robot to reach the goal from the start position. This includes teleoperation intervention times for trials that required assists.
  \item[] \textbf{Unsafe time}: The total amount of time when the robot enters into an unsafe region by crossing outside of the footpath. This was manually determined by the experimenter.
  \item[] \textbf{Number of collisions}: Number of times when the robot collided with something, plus the number of times when a collision is about to occur resulting in the experimenter having to take over control of the robot to prevent it.
  \item[] \textbf{Number of assists}:  Number of times teleoperation intervention is required, either to prevent collisions or enable the robot to continue when stuck due to LiDAR sensor noise or mislabelled footpath pixels from the semantic segmentation model blocking the robot's path.
\end{itemize} 

\section{Experimental Results}
\label{experimental-results}

%This section details the results from the indoor and outdoor tests of the three models. Our proposed model combining semantic segmentation and LiDAR had the highest success rate out of the three models indoors, with 89\% success indoors without pedestrians and 67\% with pedestrians present due to a lower number of collisions and assists than the other models. The lower success is attributed to an increased number of assists required when pedestrians were present as noise from the segmentation proved difficult for path planning in the confined space of an office. Outdoors our proposed model was more successful, again outperforming the other models. Both with and without pedestrians our model remained on the safe path of travel 100\% of the time, having a 91\% success rate without pedestrians and 100\% success rate with pedestrians. The lower success rate without pedestrians is attributed to an assist required for one trial at the differing material testing location due to noise from the segmentation preventing the robot from planning a path across the material transition. This compares 

\subsection{Outdoor Navigation}

Results for the outdoor navigation scenarios are shown in Table \ref{outdoor table}. Our proposed method performed significantly better than the two alternatives in terms of success rate, in both with and without pedestrians conditions. Specifically, our proposed method achieved above 90\% success rate when pedestrians were either present or not. On the other hand, the SS Only method achieved 66.67\% and 83.33\% respectively, and the LiDAR Only method achieved 75.00\% and 50.00\% respectively.

Examining the \textit{Unsafe Time}, \textit{Collisions}, and \textit{Assists} metrics, we see that the LiDAR Only and SS Only methods occasionally departed from the safe path of travel. Furthermore, the SS Only method had collisions and required assistance in several occasions. Using our proposed method, the robot was able to stay within the footpath for all trials, did not have any collisions, and only needed assistance on a very small number of occasions. 

\begin{table}[]
\vspace{3mm}
\caption{Outdoor navigation results}
\label{outdoor table}
\begin{adjustbox}{width=\columnwidth,center}
\centering
\begin{tabular}{@{}llllllll@{}}
\toprule
\multicolumn{2}{l}{Model}                                                             & \multicolumn{2}{l}{LiDAR Only} & \multicolumn{2}{l}{SS Only} & \multicolumn{2}{l}{Ours} \\ 
{Pedestrians}      &                                                 & No          & Yes         & No             & Yes             & No                                       & Yes                                      \\
\midrule
{Success Rate \%}                              &                     & 50.00       & 75.00       & 83.33          & 66.67           & 91.67                                    & 100.00                                   \\

\multirow{2}{*}{\begin{tabular}[c]{@{}l@{}}Trial Duration\\ (sec)\end{tabular}} & Avg & 64.18       & 62.45       & 64.01          & 81.04           & 72.76                                    & 79.05                                    \\
                                                                                & SD  & 26.30       & 25.55       & 23.77          & 33.60           & 27.76                                    & 25.21                                    \\
\multirow{2}{*}{\begin{tabular}[c]{@{}l@{}}Unsafe Time\\ (sec)\end{tabular}}    & Avg & 2.59        & 1.92        & 0.61           & 0.00            & 0.00                                     & 0.00                                     \\
                                                                                & SD  & 3.25        & 3.62        & 2.10           & 0.00            & 0.00                                     & 0.00                                     \\
\multirow{2}{*}{Collisions}                                                     & Avg & 0.00        & 0.00        & 0.00           & 0.17            & 0.00                                     & 0.00                                     \\
                                                                                & SD  & 0.00        & 0.00        & 0.00           & 0.39            & 0.00                                     & 0.00                                     \\
\multirow{2}{*}{Assists}                                                        & Avg & 0.00        & 0.00        & 0.08           & 0.25            & 0.08                                     & 0.00                                     \\
                                                                                & SD  & 0.00        & 0.00        & 0.29           & 0.45            & 0.29                                     & 0.00                                    
\\\bottomrule
\end{tabular}
\end{adjustbox}
\end{table}

\begin{figure}[b]
\centerline{\includegraphics[width=95pt]{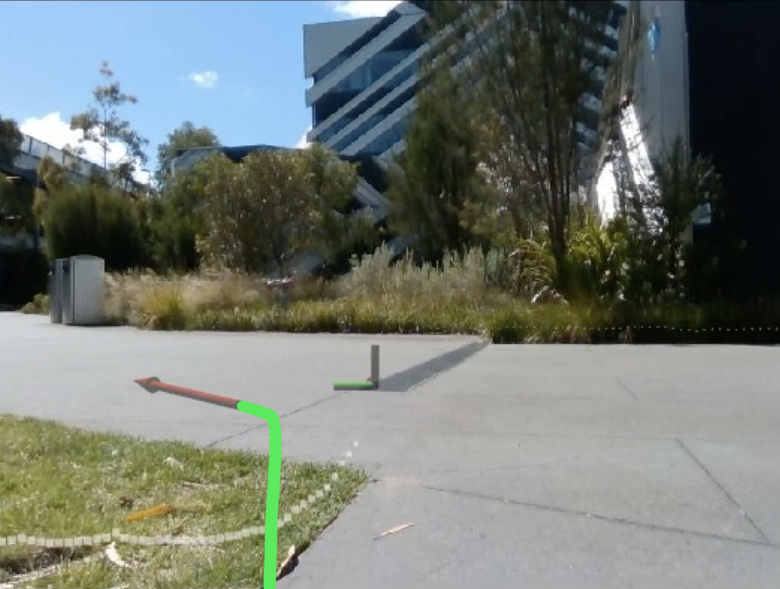} \includegraphics[width=95pt]{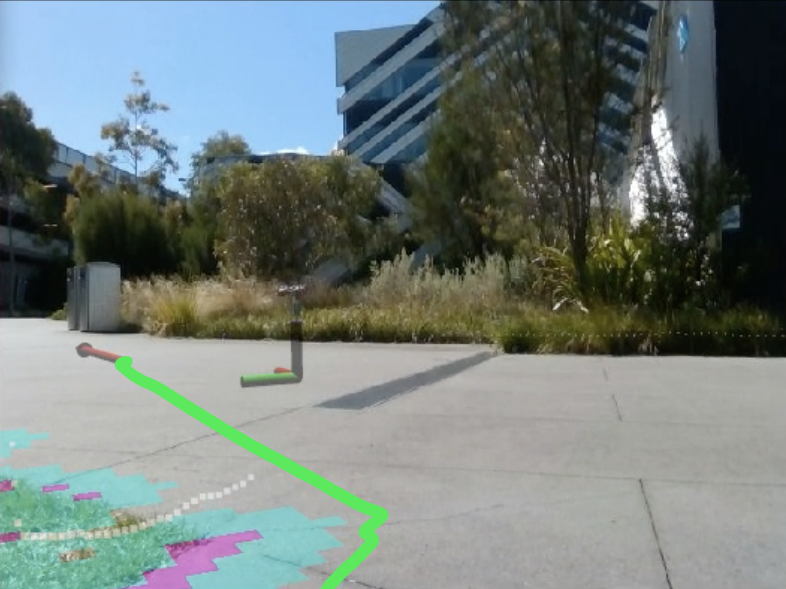}}

\centerline{\includegraphics[width=95pt]{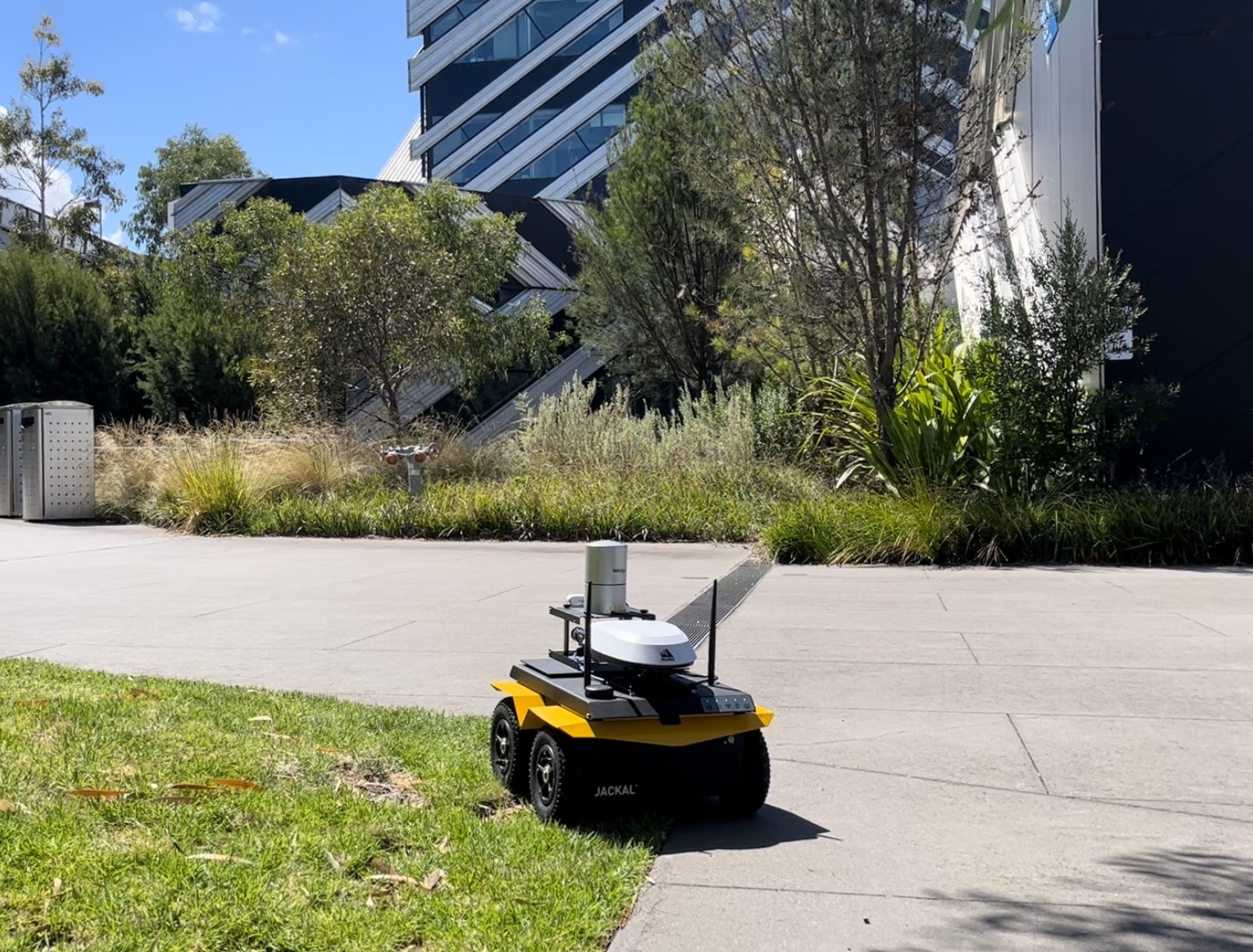} \includegraphics[width=95pt]{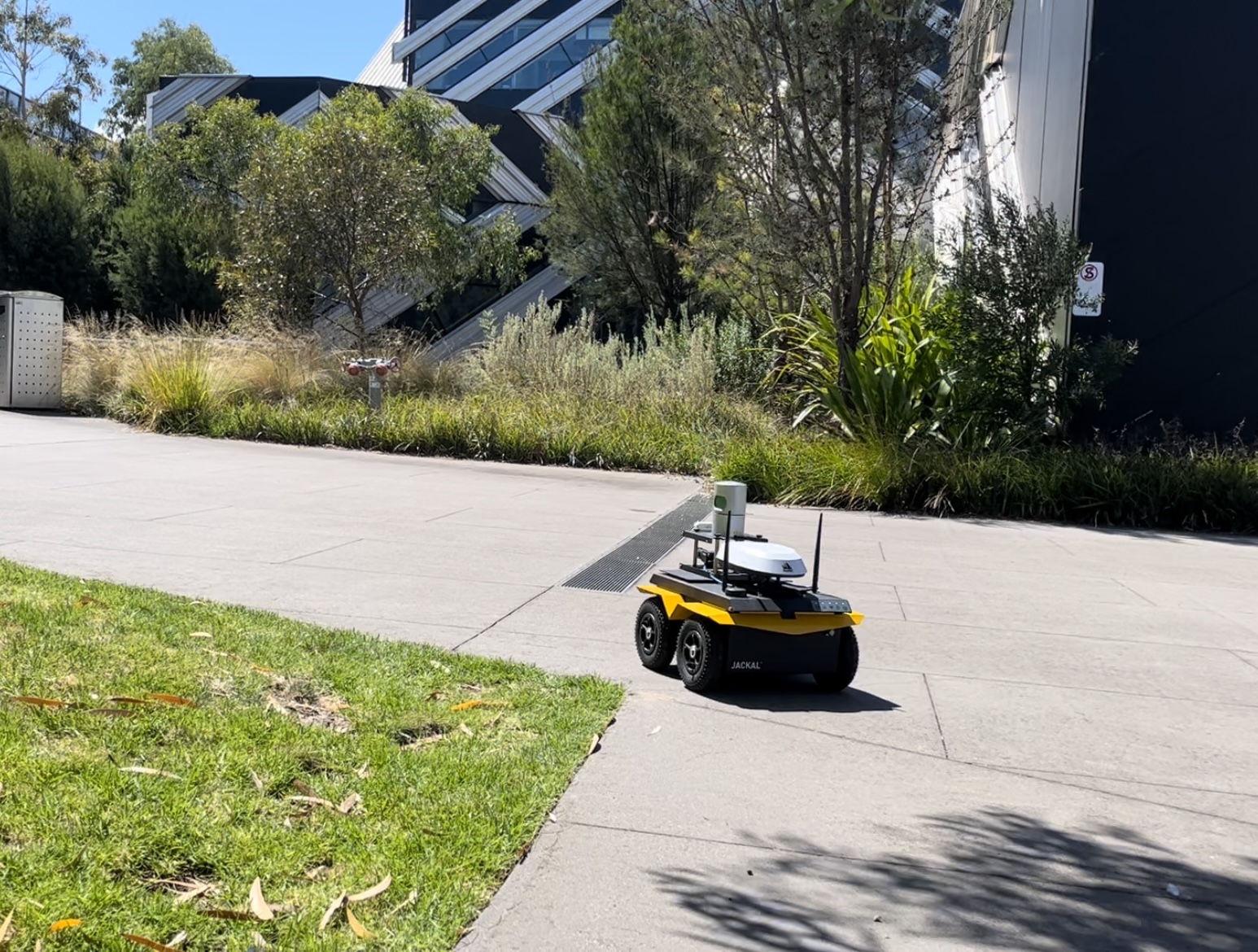}}
\caption{Impact of semantic segmentation in path planning to avoid unsafe paths. The white dots are $L_{obstacle}$, red arrow is the goal pose, green line is the path planned by the Robot Motion Planner. The pink areas represent the obstacle markings on the costmap and the blue is the inflation of these obstacles when marked on the costmap. Top left shows path planning for LiDAR Only does not avoid grass. Top right shows path planning for our model that does avoid the grass. Bottom left shows the Jackal following the path on the unsafe terrain when LiDAR Only is used. Bottom right shows the Jackal avoiding the grass to stay on the safe path when our model is used.}
\label{grass experiments}
\end{figure}

Fig.~\ref{grass experiments} shows an example scenario with grass at an outdoor location to demonstrates the performance of our method. The top left figure shows that the LiDAR Only methods plans a path that cuts across the grass areas, since the LiDAR sensor is not able to distinguish grass from footpath. The top right image shows how our method is able to detect the footpath boundary correctly and plans a path that avoids the grass area. The bottom left image shows that using the LiDAR Only method, the robot ended up travelling over the grass areas, while the bottom right images shows that using our method, the robot is able to travel through a safer and more socially conforming path by staying off the grass.

Collisions with pedestrians were the most common collision source for the segmentation model without LiDAR. Specifically this would occur when pedestrians would walk close to the robot, directly perpendicular to it's path. Pedestrians this close could not be recognised in time due to either being in the camera blind spot or blocking the entire frame leading to an incorrect segmentation mask. This was not an issue for the models that utilized LiDAR, where no collisions occurred.

\begin{figure}[b]
\centerline{\includegraphics[width=95pt]{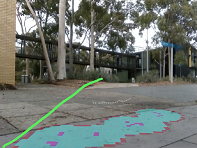} \includegraphics[width=95pt]{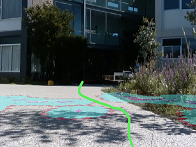}}
\caption{Path planning around mislabelled footpath pixels from semantic segmentation model. Left image shows mislabelled pixels due to the transition of materials on the path. Right image shows mislabelling due to shadows.}
\label{noise plan}
\end{figure}

\begin{figure}[b]
\centerline{\includegraphics[width=95pt]{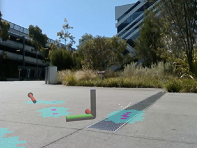} \includegraphics[width=95pt]{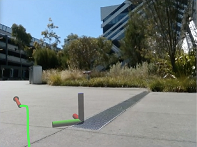} }
\caption{Semantic segmentation pixel mislabelling corrected after recovery behaviours executed. Left image shows mislabelled footpath pixels at the goal location. Right image shows this mislabelling corrected after costmap clearing and rotate recovery behaviours.}
\label{noise recovery}
\end{figure}

In environments were segmentation mislabelled footpath pixels, assistance was needed for the SS Only and the proposed model as the incorrect labelling resulted in the Robot Motion Planner being unable to calculate a valid path to the goal. This assist involved moving the robot forward to pass the mislabelled area if no path could be found after all recovery behaviours had been enacted, or to prevent the robot from a damaging collision (i.e. driving into dense garden). The outdoor environments where this mislabelling most commonly occurred were the locations with shadows and paths with material transitions. In most cases, the robot was able to plan a path around these incorrect areas as shown in Fig.~\ref{noise plan}, particularly when the path was wider as there was available space for the robot to move around. In other cases where space was unavailable, the mislabelling of pixels was corrected, either by moving the robot forward resulting in semantic segmentation model outputting a more correctly labelled mask or clearing the costmap, as shown in Fig.~\ref{noise recovery}. When an assist was required, the trial was considered to be unsuccessful. This explains why success rate for the non-pedestrian LiDAR and segmentation model was lower than for the pedestrian trials in outdoor environments as shown in Table \ref{outdoor table}.

\subsection{Indoor Navigation}
Results for the indoor navigation environments are shown in Table \ref{indoor table}. Results show that our method achieved the highest success rates among all methods. Our method had a 66.67\% success rate and 88.89\% success rate for the situations with and without pedestrians respectively. These results are at least as good as, or better than the LiDAR Only method, and significantly better than the SS Only method. Additionally, considering the \textit{Trial Durations}, our method was able to reach the goal on average 14\% faster than the LiDAR Only method and 36\% faster than the SS Only method. This is because our method allows the robot to detect obstacles more accurately, which in turn allows the robot to plan better paths that lead to fewer collisions or the robot being stuck and needing to replan its path. 

\begin{figure}[b]
\centerline{\includegraphics[width=95pt]{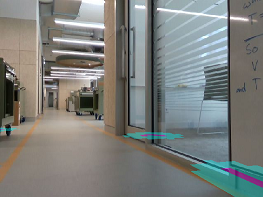} \includegraphics[width=95pt]{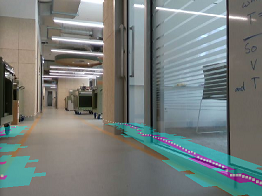}}
\caption{Semantic segmentation ability to identify glass compared to LiDAR. Left shows LiDAR only. Right shows semantic segmentation.}
\label{glass comparison}
\end{figure}

\begin{figure}[b]
\centerline{\includegraphics[width=95pt]{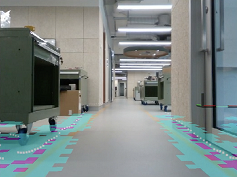} \includegraphics[width=95pt]{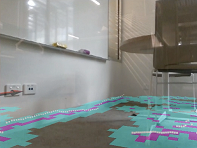}}
\caption{Problems with glass identification for segmentation. Left shows the glass door on the right being originally identified as an obstacle. Right shows glass not being identified as an obstacle as the robot gets too close to the glass door and turned to look straight through the glass.}
\label{glass collision}
\end{figure}

The \textit{Unsafe Time} for all models was 0 seconds as there were no unsafe paths in the indoor environment tested and the robot could travel safely on any parts of the floor. Considering the \textit{Collisions} and \textit{Assists} metrics, although our method resulted in some collisions and required occasional assistance, the performances are substantially better than the two alternative methods. The main collision source for the LiDAR Only method was glass doors as shown in Fig.~\ref{glass comparison}. The left image depicts how the LiDAR Only method was not able to detect the glass door as an obstacle. However, using our method, as shown in the right image, the semantic segmentation network is able to identify the glass door as a footpath boundary. The ability of our method to detect glass doors was an unexpected benefit, as we only labeled the footpath pixels in the training images, and did not explicitly label glass doors as obstacles. However, there are situations where our method would still fail to detect glass doors. This occurs when the robot gets too close to the glass door and is facing the glass door. In such cases, the robot can no longer see the edges of the glass door and instead looks through and sees the room behind instead, as shown in Fig.~\ref{glass collision}. However, these occasions are rare, and most of the time our method was able to avoid collisions with glass doors. For the SS Only method, we observed that a key source of collisions was thin table legs.

\begin{table}[]
\vspace{3mm}
\caption{Indoor navigation results}
\label{indoor table}
\begin{adjustbox}{width=\columnwidth,center}
\begin{tabular}{@{}llllllll@{}}
\toprule
\multicolumn{2}{l}{Model}                                                             & \multicolumn{2}{l}{LiDAR Only} & \multicolumn{2}{l}{SS Only} & \multicolumn{2}{l}{Ours} \\ 
{Pedestrians}                                 &                      & No          & Yes         & No              & Yes            & No                                       & Yes                                      \\
\midrule
{Success Rate \%}                             &                      & 77.78       & 66.67       & 44.44           & 52.70          & 88.89                                    & 66.67                                    \\
\multirow{2}{*}{\begin{tabular}[c]{@{}l@{}}Trial Duration\\ (sec)\end{tabular}} & Avg & 66.40       & 66.97       & 95.90           & 84.22          & 53.15                                    & 61.58                                    \\
                                                                                & SD  & 77.21       & 39.09       & 59.15           & 49.38          & 22.24                                    & 26.99                                    \\
\multirow{2}{*}{\begin{tabular}[c]{@{}l@{}}Unsafe Time\\ (sec)\end{tabular}}    & Avg & 0.00        & 0.00        & 0.00            & 0.00           & 0.00                                     & 0.00                                     \\
                                                                                & SD  & 0.00        & 0.00        & 0.00            & 0.00           & 0.00                                     & 0.00                                     \\
\multirow{2}{*}{Collisions}                                                     & Avg & 0.33        & 0.56        & 0.56            & 0.44           & 0.11                                     & 0.11                                     \\
                                                                                & SD  & 1.00        & 1.01        & 0.73            & 0.53           & 0.33                                     & 0.33                                     \\
\multirow{2}{*}{Assists}                                                        & Avg & 0.33        & 0.56        & 0.78            & 0.44           & 0.11                                     & 0.33                                     \\
                                                                                & SD  & 0.71        & 1.01        & 1.09            & 0.53           & 0.33                                     & 0.50                                    
\\\bottomrule
\end{tabular}
\end{adjustbox}
\end{table}

%A key cause of collision for the segmentation only model was chair legs. Assists were required in these cases to avoid collision and move past the chair leg so the robot could continue to attempt to reach the goal. All models had instances where the robot would collide and drive over a flat table leg on the ground.

%Table \ref{indoor table} shows that the LiDAR was more successful than the segmentation only model for indoor tests. This is likely due to the fact that the safe paths of travel in these indoor office environments are dictated by static vertical obstacles (tables, chairs, walls, etc.) that are well identified by LiDAR.

\section{Discussion}
\label{discussion}
The results above show how our proposed approach was successful in addressing the two key challenges of remaining on a safe path of travel while simultaneously avoiding static and dynamic obstacles along the way. The proposed combination of LiDAR and semantic segmentation with the ROS Navigation Stack led to a higher success rate in outdoor settings, with and without pedestrians when compared to other models. While the proposed model was less successful in indoor settings due to increased collisions and operator assistance, our model still outperformed the LiDAR Only and SS Only models in this case. The semantic segmentation model was able to address the LiDAR's inability to distinguish between safe and unsafe paths of travel beyond obstacle detection. The LiDAR was able to address the limitations of the semantic segmentation model in identifying fast approaching obstacles in close range as well as other unfamiliar obstacles along the way. The semantic segmentation model and subsequent point cloud production proved to be fast enough to safely navigate in real time. There were difficulties in real-time obstacle detection when sharply turning corners due to the limited field of view of the camera. This was solved by limiting the rotational velocity of the robot in our implementation to allow time for the model to process the new camera frames while still maintaining a suitable speed of movement.

One issue encountered by our system is related to drift in the robot's odometry. Drift error in odometry could result in a misalignment between the local costmap and the physical world. Similarly, drift error could lead the robot to a location that is off from the intended goal position. In some cases, this can result in the robot trying to navigate to a goal position that is not even reachable (e.g., behind a wall). Integrating an additional localization module could help address these issues. 

The system we have proposed is a local navigation method that allows a robot to plan short navigation paths. The viability of this approach over longer journeys would need to be investigated further, particularly due to the odometry drift issues. Due to the absence of a map, our method can also run into difficulties in path planning when there are dead ends. However, this is a common problem for local navigation methods. Our proposed method can potentially be integrated with a global navigation or planning method to enable robust navigation over longer distances.

The main cause of failure cases in indoors scenarios was the inability to see glass doors. When pedestrians were present, the narrow corridor made it difficult for the robot to move straight along the corridor, and it would attempt to turn left or right and drive through glass doors when a pedestrian was blocking the corridor. Experiment results show that our proposed method was able address this issue as the semantic segmentation model identifies glass walls/doors as obstacles and thus was able achieve a much higher success rate than the LiDAR only model.

\section{CONCLUSION AND FUTURE WORK}\label{conclusion}
In this paper we present an autonomous navigation approach for mobile robots in unknown urban environments using semantic segmentation to identify safe paths of travel combined with LiDAR and a local navigation algorithm. Using the semantic segmentation mask to create a pointcloud of the obstacle environment as well as a laser scan of path boundaries, we produced a model that successfully navigates over short distances in indoor and outdoor environments while staying on the safe path of travel. Our model was evaluated against two other models - one using only semantic segmentation and one only using LiDAR. Our evaluation shows that our model was able to match or outperform the two other models with a success rate over 90\% with and without pedestrians compared to 66.67\% and 83.33\% for the semantic segmentation only model and 75.00\% and 50.00\% for the LiDAR only model, when pedestrians were present and not present respectively in outdoor settings. In indoor settings, our model had a 66.67\% success rate  when pedestrians were present, matching the LiDAR only and outperforming the semantic segmentation only (52.70\%). Without pedestrians, our model had the highest success rate (88.89\%) compared to 77.78\% and 44.44\% for LiDAR only and segmentation only, respectively.

Further work is required on the semantic segmentation model to improve the model's ability to recognise footpaths in shaded environments and material transitions in order to achieve better performance. Also, the performance of other local navigation algorithms, such as SLAM, with our approach could be assessed. Future work is required to fix the odometry drift issues such as incorporating additional localisation modules.

\bibliographystyle{IEEEtran}
\bibliography{IEEEabrv,ref}

\end{document}